\documentclass[runningheads]{llncs}
\usepackage{graphicx}
\usepackage{todonotes}
\usepackage{siunitx}
\usepackage{subcaption}
\usepackage{booktabs}
\usepackage{xspace}

\usepackage{comment}
\usepackage{amsmath,amssymb}
\let\vec\mathbf
\usepackage{color}
\usepackage{url}
\usepackage{hyperref}

\newcommand{\etal}{\emph{et~al}.\xspace}
\newcommand{\wrt}{w.r.t.\xspace}
\newcommand{\pcent}[1]{\SI{#1}{\percent}}

\begin{document}



\title{Simplified Concrete Dropout - Improving the Generation of Attribution Masks for Fine-grained Classification}
\titlerunning{Simplified Concrete Dropout - Improving Attribution Masks for FGVC}

\author{
Dimitri Korsch\inst{1} \orcidID{0000-0001-7187-1151}
\and
Maha Shadaydeh\inst{1} \orcidID{0000-0001-6455-2400}
\and
Joachim Denzler\inst{1} \orcidID{0000-0002-3193-3300}
}
\authorrunning{D. Korsch et al.}
%
\institute{
Computer Vision Group, Friedrich Schiller University Jena, Jena, Germany\\
\email{\{dimitri.korsch,maha.shadaydeh,joachim.denzler\}@uni-jena.de}
\\
\url{https://inf-cv.uni-jena.de}
}

\maketitle

\begin{abstract}
	Fine-grained classification is a particular case of a classification problem, aiming to classify objects that share the visual appearance and can only be distinguished by subtle differences.
	Fine-grained classification models are often deployed to determine animal species or individuals in automated animal monitoring systems.
	Precise visual explanations of the model's decision are crucial to analyze systematic errors.
	Attention- or gradient-based methods are commonly used to identify regions in the image that contribute the most to the classification decision.
	These methods deliver either too coarse or too noisy explanations, unsuitable for identifying subtle visual differences reliably.
	However, perturbation-based methods can precisely identify pixels causally responsible for the classification result.
	\emph{Fill-in of the dropout} (FIDO) algorithm is one of those methods.
	It utilizes the \emph{concrete dropout} (CD) to sample a set of attribution masks and updates the sampling parameters based on the output of the classification model.
	A known problem of the algorithm is a high variance in the gradient estimates, which the authors have mitigated until now by mini-batch updates of the sampling parameters.
	This paper presents a solution to circumvent these computational instabilities by simplifying the CD sampling and reducing reliance on large mini-batch sizes.
	First, it allows estimating the parameters with smaller mini-batch sizes without losing the quality of the estimates but with a reduced computational effort.
	Furthermore, our solution produces finer and more coherent attribution masks.
	Finally, we use the resulting attribution masks to improve the classification performance of a trained model without additional fine-tuning of the model.
\end{abstract}

\section{Introduction}
\label{sec:intro}

Fine-grained classification tackles the hard task of classifying objects that share the visual appearance and can only be distinguished by subtle differences, e.g., animal species or car makes.
Most commonly, fine-grained classification models are employed in the field of animal species recognition or animal individual identification: classification of insects~\cite{bjerge2021automated,Korsch22:AVM_NID} and birds~\cite{hu2019see,Korsch21:ETE,Simon19:Implicit}, or identification of elephants~\cite{Koerschens19:ELPephants}, great apes~\cite{Brust2017AVM,Kaeding18_ALR,sakib2021visual,yang2019great}, and sharks~\cite{hughes2017automated}.
Even though these automated recognition systems surpass humans in terms of recognition performance, in some cases, an explanation of the system's decision might be beneficial even for experts.
On the one hand, explanations might help in cases of uncertainty in human decisions.
On the other hand, it can help to feedback information to the developer of the system if systematic errors in the decision are observable.
Those systematic errors might be spurious biases in the learned models~\cite{reimers2021conditional} and could be revealed by inspection of a highlighted region that should not be considered by a classification model.

Even though various methods \cite{hu2019see,Korsch21:ETE,Koerschens19:ELPephants,Simon19:Implicit} were presented in the context of fine-grained recognition to reliably distinguish classes with subtle visual differences, these methods offer either a too coarse-grained visual explanation or an explanation with many false positives.

Attention-based methods~\cite{he2019and,hu2019see,zhang2019learning}, for example, introduce attention mechanisms to enhance or diminish the values of intermediate features.
They operate on intermediate feature representations, which always have a much lower resolution than the input image.
Hence, upscaling the low-resolution attention to the higher-resolution image cannot highlight the fine-grained areas, which are often important for a reliable explanation of the decision.

Gradient-based methods~\cite{simonyan2014deep,guided_backprop,integrated_grads} identify pixel-wise importance by computing the gradients of the classifier's decision \wrt the input image.
These methods identify much finer areas in the image and enable decision visualization on the fine-grained level.
However, these methods may also falsely highlight background pixels as has been shown in the work of Shrikumar~\etal~\cite{shrikumar2017} and Adebayo~\etal~\cite{adebayo2018}.

In this paper, we build upon a perturbation-based method, the
fill-in of the dropout (FIDO) approach, proposed by Chang~\etal~\cite{chang2018explaining}.
The idea behind FIDO is to perturb the pixel values of the input image and observe the change in the classification decision.
The authors realize the perturbation with a binary mask, whose entries model a binary decision whether to perturb a pixel or not.
The mask is sampled using a set of trainable parameters and a sampling method introduced as concrete dropout (CD) by Gal~\etal~\cite{gal2017concrete}.
After optimizing the trainable parameters \wrt the classification decision, the parameters represent the importance of each pixel for the classification.
One drawback of the approach is the high variance in estimating the gradients while optimizing the sampling parameters.
Chang~\etal mention this drawback in their work, and suggest reducing the variance by averaged gradients over a mini-batch of sampled masks.

\begin{figure}[t]
	\includegraphics[width=\textwidth]{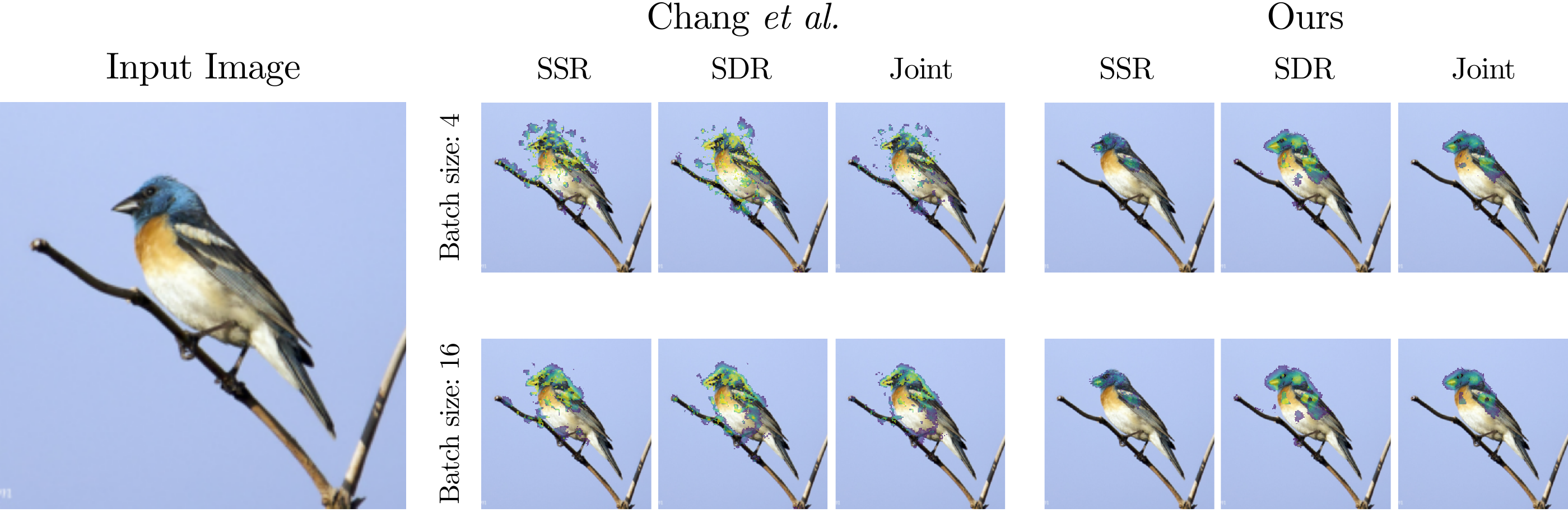}
	\caption{Visual comparison of the original FIDO method and our proposed improvement. We estimated the masks using \num{30} optimization steps and two different batch sizes: \num{4} and \num{16}. Our method produces masks that do not differ much for the visualized batch sizes. In contrast, the method of Chang~\etal~\cite{chang2018explaining} strongly depends on higher batch sizes and produces more wrongly attributed pixels (e.g.,~the background or the tree branch) if the mini-batch size of \num{4} or less is used. Similarly to Chang~\emph{et al.}, we visualized only the mask values above the threshold of \num{0.5}. \emph{(best viewed in color)}}
	\label{fig:example_images}
\end{figure}

We propose a mathematically equivalent but simplified version of CD.
As a consequence, we reduce the amount of exponential and logarithm operations during the sampling procedure resulting in more stable gradient computations.
We will show that the FIDO algorithm becomes less reliant on the size of the mini-batches, which allows the estimation of the attribution masks with smaller mini-batch sizes.
To summarize our contribution:
(1) the estimation of the attribution masks is possible with a smaller mini-batch size without losing quality but with a reduced computational effort;
(2) our proposed solution results in more precise and coherent attribution masks, as Figure~\ref{fig:example_images} shows;
(3) most importantly, we demonstrate that the estimated attribution mask can be leveraged for improving the classification performance of a fine-grained classification model.
We outperform other baseline methods and achieve classification results comparable to a setup if ground truth bounding boxes are used.

\section{Related Work}
\label{sec:related}

\noindent\textbf{Attribution methods} aim at estimating a saliency (or attribution) map for an input image that defines the importance of each area in the image for the desired task, e.g., for classification.
We give a brief overview of three possible attribution methods that are often used in literature.

End-to-end trainable \emph{attention methods}~\cite{he2019and,hu2019see,zhang2019learning} present different approaches that modify the architecture of a CNN model.
In general, these modifications estimate saliency maps, or attentions, for intermediate feature representations.
The estimated saliencies are then used to enhance or diminish the feature values.
Even though the estimated attention maps improve the classification significantly, they are coarse since they typically operate on intermediate representations right before the final classification layer.
Consequently, up-scaling these attentions to the dimension of the original image cannot capture precisely the fine-grained details.

In contrast, \emph{gradient-based methods}~\cite{Korsch19_CSPARTS,simonyan2014deep,guided_backprop,integrated_grads} estimate pixel-wise importance by computing gradients of the outputs (logit of the target class) w.r.t. the input pixels.
Even though the resulting saliency map is much finer than attention-based saliencies, it often highlights lots of irrelevant areas in the image, e.g., the background of the image.
This may be caused by gradient saturation, discontinuity in the activations of the network~\cite{shrikumar2017}, or an inductive bias due to the convolutional architecture, which is independent of the learned parameters~\cite{adebayo2018}.

Finally, \emph{perturbation-based methods}~\cite{chang2018explaining,dabkowski2017real,fong2017interpretable} attribute the importance to a pixel by modifying the pixel's value and observing the change in the network's output.
These methods identify image regions that are significantly relevant for a given classification target.
Hereby, two different objectives can be used to estimate these regions:
(1) the estimation of the smallest region that retains a certain classification score, and (2) the estimation of the smallest region that minimizes the target classification score when this region is changed.
Dabkowski and Gal~\cite{dabkowski2017real} presented a method that follows the mentioned objectives and Chang~\etal~\cite{chang2018explaining} reformulated these objectives in their fill-in of the dropout (FIDO) algorithm.
They further presented different infill methods and their effects on the estimated saliency maps.

In this work, we utilize the FIDO algorithm and present a way to enhance the computation stability of the gradients.
Furthermore, we propose a way to combine the resulting attribution masks into a joint attribution mask, which we finally use to improve the results of a fine-tuned classification model.
As with all perturbation-based methods, we keep the advantage that neither a change of the architecture nor a fine-tuning of the parameters is required.

\noindent\textbf{Fine-grained categorization} is a special classification discipline that aims at distinguishing visually very similar objects, e.g., bird species~\cite{WahCUB_200_2011}, car models~\cite{StanfordCars}, moth species~\cite{Rodner15:FRD}, or elephant individuals~\cite{Koerschens19:ELPephants}.
These objects often differ only in subtle visual features and the major challenge is to build a classification model that identifies these features reliably.
On the one hand, it is common to utilize the input image as it is and either perform a smart \emph{pre-training strategy}~\cite{Cui_2018_CVPR_large,krause2016unreasonable} or \emph{aggregate feature} using different techniques~\cite{lin2015bilinear,Simon19:Implicit}.
On the other hand, there are the \emph{part- or attention-based approaches}~\cite{he2019and,hu2019see,Korsch19_CSPARTS,zhang2019learning} that either extract relevant regions, so-called parts, in the input image or enhance and diminish intermediate feature values with attention mechanisms.
Finally, \emph{transformer-based approaches}~\cite{he2022transfg,yu2022metaformer} are currently at the top in different fine-grained classification benchmarks.
However, these methods rely on the parameter-rich transformer architecture and big datasets for fine-tuning.
Typically, in the context of automated animal monitoring these resources are not available making such models difficult to deploy in the field.

In this work, we use two widely used CNN architectures~\cite{resnet,inceptionv3} fine-tuned on the CUB-200-2011~\cite{WahCUB_200_2011} dataset.
Without any further modification or fine-tuning, we use the estimated attribution masks to extract an auxiliary crop of the original input image.
Finally, we use the extracted crop to enhance the classification decision.

\section{Simplified Concrete Dropout - Improved Stability }
\label{sec:methods}
Our final goal is a pixel-wise attribution of importance for a certain classification output.
Perturbation-based methods offer a way to estimate this importance by observing the causal relation between a perturbation of the input and the caused change in the decision of a classification model.
One example is the \emph{fill-in of the dropout} (FIDO) algorithm~\cite{chang2018explaining} which computes these attribution masks by identifying pixel regions defined as following:

\begin{enumerate}
	\item Smallest destroying region (SDR) represents an image region that \emph{minimizes} the classification score if this region is changed.
	\item Smallest sufficient region (SSR) represents an image region that \emph{maximizes} the classification score if only this region is retained from the original content.
\end{enumerate}
Based on these definitions, the FIDO algorithm optimizes the parameters of a binary dropout mask.
The mask identifies whether a pixel value should be perturbed with an alternative representation (\emph{infill}) or not, and can be interpreted as a saliency map.
In the following, we formalize the objective functions, illustrate the limitation of the FIDO algorithm, and explain our suggested improvements.

\subsection{The FIDO Algorithm and Its Limitations}
\label{sub:problem_formulation}

Given an image $\vec{x}$ with $N$ pixels, a class $c$, a differential classification model $\mathcal{M}$ producing an output distribution $p_{\mathcal{M}}(c|\vec{x})$, we are interested in a subset of pixels $r$ that divides the image into two parts $\vec{x}=\vec{x}_r \cup \vec{x}_{\setminus r}$.
Observing the classifier's output when $\vec{x}_r$ is not visible gives insights into the importance of the region $r$ for the classification decision.
Because of the binary division of the image into two parts, the region $r$ can be modeled by a binary dropout mask $\vec{z} \in \{0, 1\}^N$  with the same size\footnote{for sake of simplicity, we consider $\vec{x}$ and $\vec{z}$ as 1D vectors, instead of 2D matrices with dimensions $H$ and $W$, and $N=H \cdot W.$} as $\vec{x}$ and an infill function $\phi$ that linearly combines the original image $\vec{x}$ and an infill~$\vec{\hat{x}}$ with an element-wise multiplication $\odot$:

\begin{align}
\phi(\vec{x}, \vec{z}) &=
	(\vec{1} - \vec{z}) \odot \vec{x} + \vec{z} \odot \vec{\hat{x}} \quad.
	\label{eq:infill_function}
\end{align}

There are different ways to generate an infill image $\vec{\hat{x}}$.
First, content-independent approaches, like random (uniformly or normally distributed), or fixed (e.g., zeros) pixel values, generate the infill image independently of the input's content.
Because these methods are independent of the content of the image, they cause a hard domain shift for the underlying classification model.
Chang~\etal~\cite{chang2018explaining} showed that these infill approaches perform worse compared to content-aware methods like GANs~\cite{gan} or Gaussian blur.
Popescu~\etal~\cite{Popescu21Knockoffs} used knockoffs~\cite{knockoffs} to generate infill values and reported in their work superiority of knockoff infills on the MNIST dataset~\cite{mnist}.
All of these content-aware methods generate the infill image depending on the pixels of the original image and retain the structure and composition of the original image to some degree.
In our experiments, we use the Gaussian blur approach to create the infill image $\hat{\vec{x}}$.
First, it removes the fine-grained details we aim to identify but retains the contents of the image so that the infill is not an out-of-domain input for the classification model.
Second, the knockoff generation, proposed by Popescu~\etal, is suitable for MNIST images because of the low dimensionality of the images ($28\times28$ px) and the binary pixel values.
Unfortunately, to this point, there is no way to apply this generation process to real-world RGB images.
Finally, the GAN-based infills are computationally intensive and require an additional model that has to be trained on data related to the images we want to analyze.

The search space of all possible binary masks $\vec{z}$ grows exponentially with the number of pixels, hence we need an efficient way to estimate the values $z_n \in \vec{z}$.
Assuming a Bernoulli distribution for the binary mask values allows us to sample the masks from a parametrized distribution $q_\theta(\vec{z})$ and optimize the parameters $\theta_n \in \theta$ using the SSR and SDR objectives:

\begin{align}
L_{SSR}(\theta) &=
	\mathbb{E}_{q_\theta(\vec{z})} \left[- s_{\mathcal{M}}(c|\phi(\vec{x}, \vec{z})) + \lambda||\vec{1} - \vec{z}||_1 \right]
\quad \text{and}\\
L_{SDR}(\theta) &=
	\mathbb{E}_{q_\theta(\vec{z})} \left[\phantom{-}s_{\mathcal{M}}(c|\phi(\vec{x}, \vec{z})) + \lambda||\vec{z}||_1 \right]
\end{align}
with the $L_1$ regularization factor $\lambda$.
Following the original work, we set $\lambda=\num{0.001}$.
The score $s_{\mathcal{M}}$ is defined as log-odds of classification probabilities:

\begin{align}
s_{\mathcal{M}}(c|\vec{x}) &= \log \frac{p_{\mathcal{M}}(c|\vec{x})}{1-p_{\mathcal{M}}(c|\vec{x})} \quad.
\end{align}

To be able to optimize $\theta$ through a discrete random mask $\vec{z}$, the authors relax the discrete Bernoulli distribution and replace it with a continuous approximation: the concrete distribution~\cite{jang2016categorical,maddison2017concrete}.
The resulting sampling, called \emph{concrete dropout} (CD), was proposed by Gal~\etal\cite{gal2017concrete} and is defined as

\begin{align}
z_n &=
	\sigma \left(\frac{1}{t}\left(\log \frac{\theta_n}{1-\theta_n} + \log\frac{\eta}{1-\eta}\right)\right) \quad \eta \sim \mathcal{U}(0,1) \label{eq:concrete_dropout}
\end{align}
with the temperature parameter $t$ (we follow the original work and set this parameter to $\num{0.1}$), $\sigma$ being the sigmoid function, and $\eta$ sampled from a uniform distribution.

In the original work, Chang~\etal~\cite{chang2018explaining} proposed two methods to speed up convergence and avoid unnatural artifacts during the optimization process.
First, they computed the gradients w.r.t. $\theta$ from a mini-batch of dropout masks.
They mentioned in Appendix (A.6) that they observed unsatisfactory results with mini-batch sizes less than 4, which they attributed to the \emph{high variance} in the gradient estimates.
Second, they sampled a coarser dropout mask (e.g. $56\times56$) and upsampled the mask using bi-linear interpolation to the dimensions of the input (e.g. $224\times224$).

In the following, we reflect upon the cause of the high variance in the gradient estimates and propose a way to increase the computational stability.
Consequently, our solution reduces the dependency of the FIDO method on the mini-batch size, allowing the estimation of the attribution masks with lower mini-batch sizes which ultimately reduces the computation time.
Finally, since we are interested in fine-grained details in the image, the estimation of a coarser attribution mask followed by an upsampling operation would not lead to the desired level of detail.
With our solution and the resulting improvement in the gradient computation, we can directly estimate a full-sized attribution mask $\theta$ without any unnatural artifacts, as shown in Figure~\ref{fig:example_images}.

\subsection{Improving Computational Stability}
\label{sub:improving_stability}
The sampling of $\vec{z}$ using CD requires that all dropout parameters $\theta_n \in \theta$ are in the range $[0,1]$.
One way to achieve this is to initialize the attribution mask with real-valued parameters $\vartheta_n \in \mathbb{R}$ and apply the sigmoid function to those: $\theta_n = \sigma (\vartheta_n)$.
As a consequence, the CD sampling procedure for $\vec{z}$, as described in Eq.~\ref{eq:concrete_dropout}, is a chaining of multiple exponential and logarithmic operations.
This can be easily implemented in the current deep learning frameworks and is a common practice, e.g., in the reference implementation of Gal~\etal\footnote{\url{https://github.com/yaringal/ConcreteDropout}}.
However, we hypothesize that exactly this chaining of operations causes a high variance in the gradient estimates, and we validate this assumption in our experiments.

Under the assumption that $\theta$ is the output of the sigmoid function, we can simplify the sampling procedure of the attribution mask $\vec{z}$ and mitigate the before-mentioned problem.
First, for readability reasons, we substitute the uniform noise part with a single variable $\hat\eta = \log \frac{\eta}{1-\eta}$ in Eq.~\ref{eq:concrete_dropout}.
Then after using the transformation $\theta_n = \sigma (\vartheta_n)$ from above, expanding the argument of the sigmoid function, and simplifying the terms, the sampling of the binary mask $\vec{z}$ using CD transforms to a simple sigmoid function:

\begin{align}
	z_n &= \sigma\left(
	\frac{1}{t}
	\left(\log \frac{\sigma\left( \vartheta_n \right) }{1-\sigma\left(\vartheta_n\right)} + \hat\eta \right)
	\right) \\
	&= \sigma\left(
		\frac{1}{t}
		\left(\log \frac{\left( 1+\exp(-\vartheta_n)\right)^{-1}}{1-\left( 1+\exp(-\vartheta_n)\right)^{-1}} + \hat\eta \right)
	\right) \label{eq:simple_concrete_dropout_start} \\
	&= \sigma\left(
		\frac{1}{t}
		\left(\log \frac{1}{\exp(-\vartheta_n)} + \hat\eta \right)
	\right)
	= \sigma \left(\frac{\vartheta_n + \hat\eta}{t}\right) \quad .
	\label{eq:simple_concrete_dropout}
\end{align}
The resulting formula is equivalent to the original formulation of CD.
However, the reduction of exponential and logarithmic operations, and hence the reduction of the number of operations in the gradient computation, are the major benefits of the simplified version in Eq.~\ref{eq:simple_concrete_dropout}.
As a result, it reduces computational inaccuracies and enhances the propagation of the gradients.
Consequently, the optimization of the parameters $\vartheta$, and of the attribution map defined by $\theta$, converges to more precise and better results as we show in Section~\ref{sec:experiments}.

In Sect.~S2 of our supplementary material, we performed an empirical evaluation of this statement and showed that our proposed simplifications result in a lower variance of the gradient estimates.
Additionally, you can find in Sect.~S1 a Python implementation of the improved Concrete Dropout layer using the PyTorch~\cite{pytorch} framework.

\subsection{Combined Attribution Mask for Fine-grained Classification}
\label{sub:combining_the_objectives}
So far, we only applied operations on the original formulation to simplify Eq.~\ref{eq:concrete_dropout} from the original work of Chang~\etal~\cite{chang2018explaining}.
However, we can also show that the estimated attribution masks improve the performance of a classification model.
First, we follow Chang~\etal and only consider mask entries with an importance rate above \num{0.5}.
Then, we estimate a bounding box around the selected values of the attribution mask.
In the end, we use this bounding box to crop a patch from the input image and use it as an additional input to the classification model (see Section~\ref{sub:improving_classification}).
Instead of using the attribution masks separately, we are interested in regions that are important to sustain and should not be deleted.
Hence, we propose to combine the attribution masks $\theta_{SSR}$ and $\theta_{SDR}$ using element-wise multiplication of mask values followed by a square root as a normalization function:

\begin{align}
	\theta_{joint} = \sqrt{\theta_{SSR} \odot (\vec{1} - \theta_{SDR} )} \quad .
	\label{eq:joint_mask}
\end{align}

With element-wise multiplication, we ensure that if either of the attribution values is low, then the joint attribution is also low.
The square root normalizes the joint values to the range where we can apply the same threshold (\num{0.5}): for example, if both attribution values are around \num{0.5}, then also the joint attribution will be around \num{0.5} and not around \num{0.25}.

\section{Experiments}
\label{sec:experiments}

We performed all experiments on the CUB-200-2011~\cite{WahCUB_200_2011} dataset.
It consists of \num{5994} training and \num{5794} test images for \num{200} different species of birds.
It is the most used fine-grained dataset for benchmarking because of its balanced sample distribution.
We selected this dataset mainly because it also contains ground truth bounding box annotations, which we used as one of our baselines in Section~\ref{sub:improving_classification}

Figure~\ref{fig:example_images} shows a qualitative comparison of the estimated masks on one example image of the CUB-200-2011 dataset.
Notably, our solution is more stable when we use smaller mini-batch sizes and produces fewer false positive attributions, e.g., in the background or highlighting of the tree branch.

We evaluated two widely used CNN architectures pre-trained on the ImageNet~\cite{ImageNet} dataset: ResNet50~\cite{resnet} and InceptionV3~\cite{inceptionv3}.
Additionally, we used an alternative pre-training on the iNaturalist2017 dataset~\cite{iNaturalist} for the InceptionV3 architecture proposed by Cui~\etal~\cite{Cui_2018_CVPR_large} (denoted with \textsc{IncV3*} and \textsc{InceptionV3*} in Tables~\ref{tab:tv_evalutation} and \ref{tab:clf_evalutation}, respectively).
All architectures are fine-tuned for \num{60} epochs on the CUB-200-2011 dataset using the AdamW~\cite{adamw} optimizer with the learning rate of \num{1e-3} (and $\epsilon$ set to \num{0.1})\footnote{Changing the default parameter smooths the training, as suggested in \\ \scriptsize{\url{https://www.tensorflow.org/api_docs/python/tf/keras/optimizers/Adam\#notes_2}}}.

\subsection{Evaluating Mask Precision}
\label{sub:exp:comparison}

To quantify the visual observations in Figure~\ref{fig:example_images}, we selected two visually similar classes of Blackbirds from the CUB-200-2011 dataset: the Red-winged Blackbird and the Yellow-headed Blackbird.
Both belong to the family of Icterids (New World blackbirds) and have black as a dominant plumage color.
However, as the name of the species indicates, the main visual feature distinguishing these birds from other black-feathered birds is the \textbf{red wing} or the \textbf{yellow head}.
Using this information, we created segmentation masks with the \emph{Segment Anything} model~\cite{sam} for the mentioned regions and used these as ground truth.

\begin{figure}[t]
	\centering
	\includegraphics[width=\textwidth]{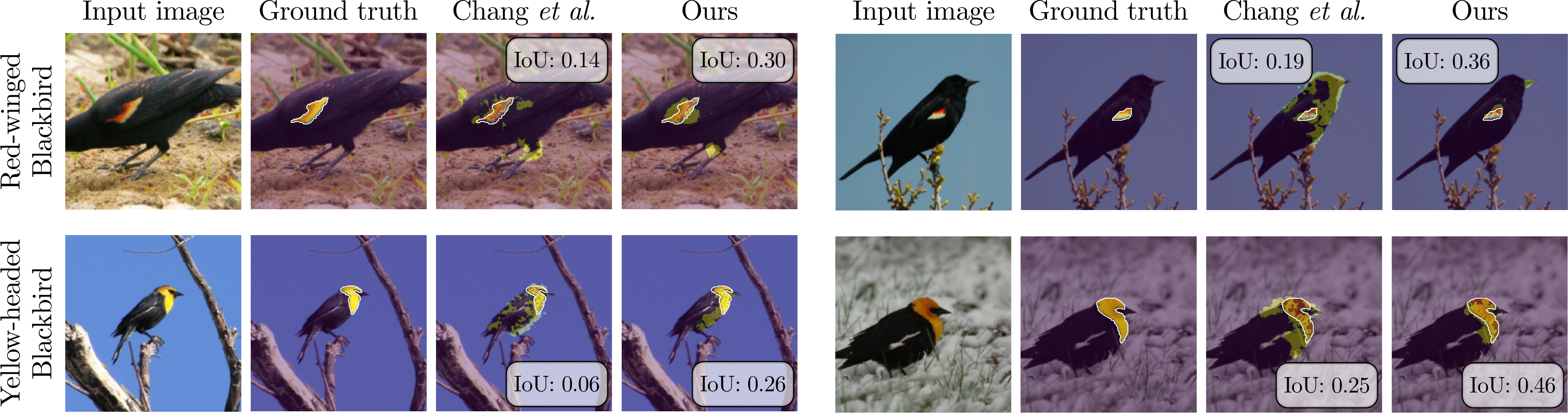}
	\caption{Examples of the estimated masks for the \emph{Red-winged Blackbird} and the \emph{Yellow-headed Blackbird} using the original FIDO approach and our proposed improved method. Besides the ground-truth segmentation masks, we also reported the intersection over union (IoU) of the estimated mask with the ground-truth.}
	\label{fig:seg_examples}
\end{figure}

We fine-tuned a classification model (ResNet50~\cite{resnet}) on the entire dataset and estimated the attribution masks using both methods: the original FIDO approach by Chang~\etal\cite{chang2018explaining} and our improved method.
We evaluated different mini-batch sizes and a different number of optimization steps.
Both of these parameters strongly affect the runtime of the algorithm.
After an attribution mask was estimated, we selected only the values above the threshold of \num{0.5} and computed the IoU with the ground truth mask.
We performed this evaluation for the masks estimated by the SSR and SDR objectives separately as well as using the joint mask as defined in Eq.~\ref{eq:joint_mask}.

\begin{figure}[t]
	\centering
	\begin{subfigure}[t]{\textwidth}
		\includegraphics[width=\textwidth]{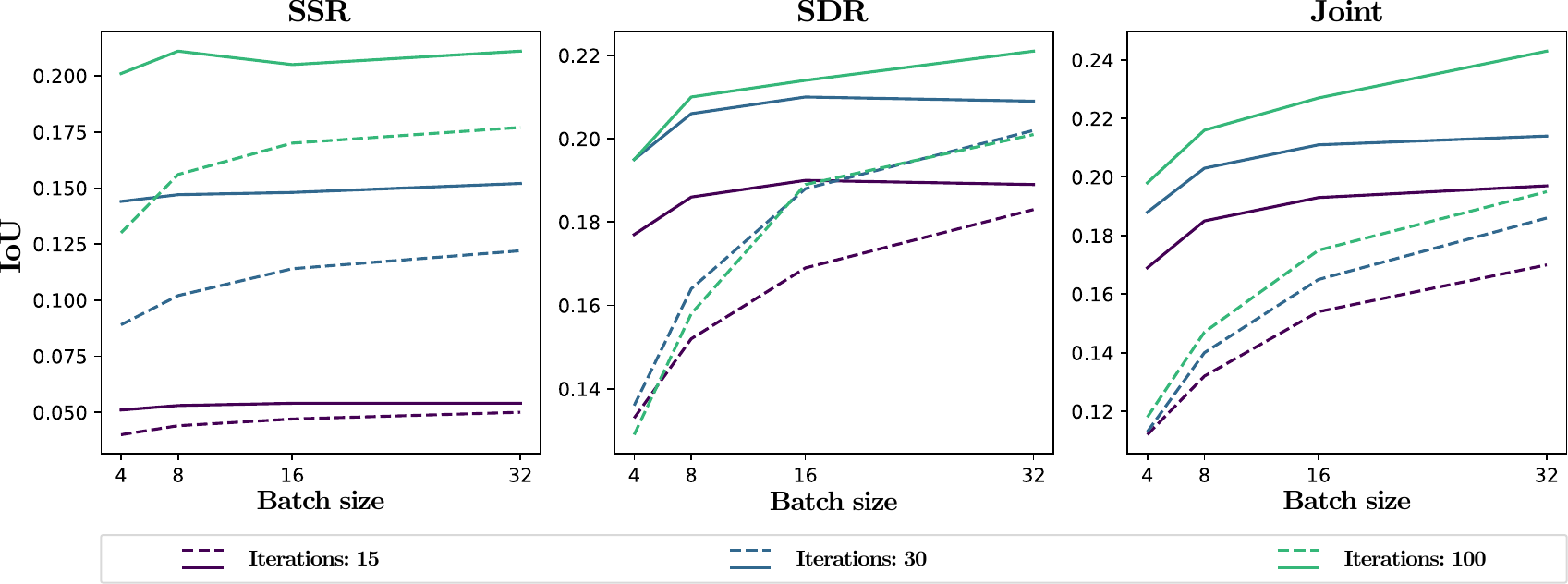}
		\caption{IoU values for different mini-batch sizes.}
		\label{fig:results:seg_by_BS}
	\end{subfigure}
	\begin{subfigure}[t]{\textwidth}
		\includegraphics[width=\textwidth]{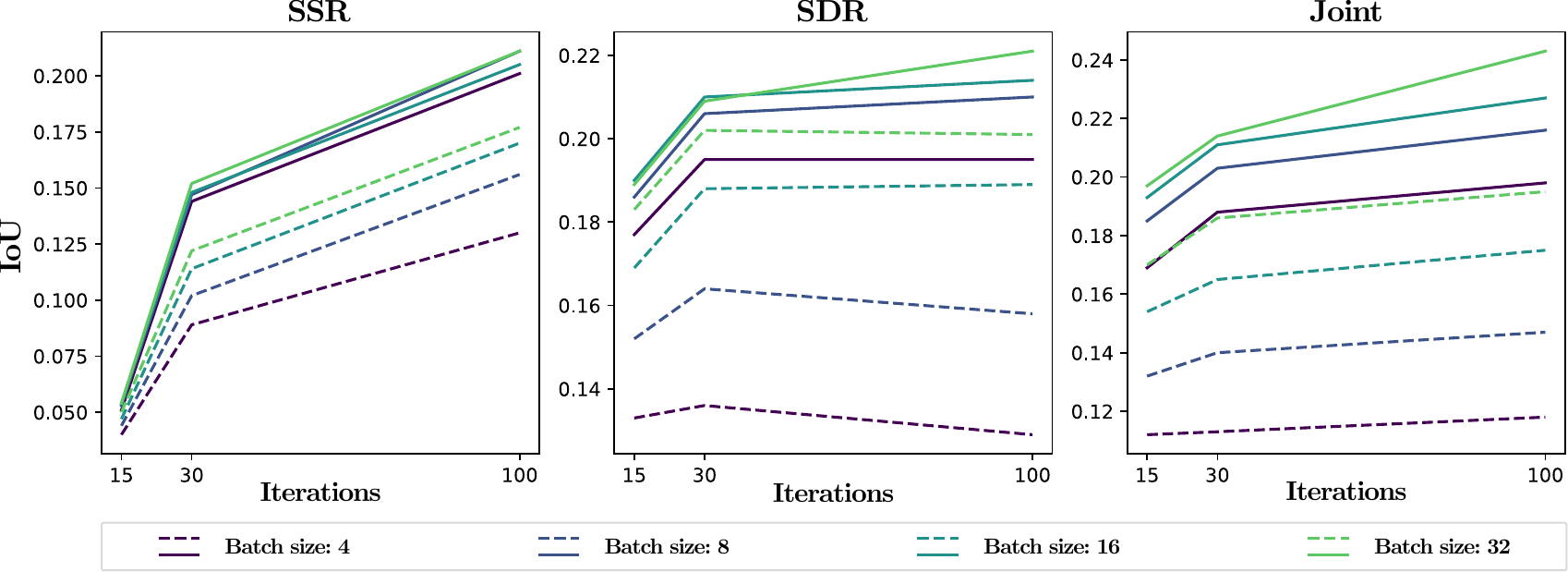}
		\caption{IoU values for a different number of optimization steps.}
		\label{fig:results:seg_by_Iters}
	\end{subfigure}
	\caption{Intersection over union (IoU) of the estimated masks with the ground-truth annotations of a discriminative region. We tested different values for the hyperparameters mini-batch size (a) and the number of optimization steps (b). Our proposed method (solid lines) outperforms the original work (dashed lines) and shows less sensitivity against the mini-batch size. \emph{(best viewed in color)}}
	\label{fig:results:segmentations}
\end{figure}

In Figure~\ref{fig:results:segmentations}, we report the IoU results of the mentioned setups.
First, the plot shows that our solution (solid lines) outperforms the original approach (dashed lines) in every constellation of the hyperparameters.
Next, the optimization process becomes less sensitive to the size of the mini-batches.
This can be seen either by the slope (Figure~\ref{fig:results:seg_by_BS}) or the variance (Figure~\ref{fig:results:seg_by_Iters}) of the IoU curves.
Our method achieved the same quality of the attribution masks with smaller mini-batch sizes.
Consequently, by reducing the mini-batch size, the number of sampled dropout masks at every optimization step is also reduced.
Hence, by using a mini-batch size of \num{8} instead of \num{32}, which Chang~\etal use in their work, we could reduce the computation time per image from \num{40} to \num{11} seconds\footnote{We processed the images using an Intel i9-10940X CPU, 128GB RAM, and a GeForce RTX 3090 GPU} for \num{100} optimization steps.

In Figure~\ref{fig:seg_examples}, we visualized four examples of the mentioned classes, our annotated segmentation masks, and the results of both approaches after \num{100} optimization steps and using a mini-batch size of \num{8}.

\subsection{Mask Coherency}
\label{sub:mask_coherency}

Following Dabkowski~\etal~\cite{dabkowski2017real}, Chang~\etal propose to use total variation regularization with a weighting factor of \num{0.01}, which is defined as

\begin{align}
	\text{TV}(\vec{z}) &= \sum_{i,j}{\left(z_{i,j} - z_{i,j+1}\right)^2} + \sum_{i,j}{\left(z_{i,j} - z_{i+1,j}\right)^2} \quad .
	\label{eq:total_variation}
\end{align}
We observed that the high variance in the gradients affects the coherency of the masks (see Figure~\ref{fig:example_images}).
Hence, we computed the total variance of the estimated masks and reported the results in Table~\ref{tab:tv_evalutation}.
The total variation is computed for the attribution masks estimated for the entire CUB-200-2011 dataset with the original approach and our proposed solution.
The results show that our solution produces more coherent masks, meaning the identified regions are more connected.

\begin{table}[t]
	\caption{Comparison of the original solution proposed by Chang~\etal and our improved implementation in terms of mask coherency. For different model architectures and different mask estimation objectives, we report the total variation as defined in Eq.~\ref{eq:total_variation} (\emph{lower is better}).}
	\centering
	\begin{tabular}{
		l@{\hspace{1.5em}}
		c@{\hspace{0.8em}}c@{\hspace{0.8em}}c@{\hspace{0.8em}}
		c@{\hspace{0.8em}}c@{\hspace{0.8em}}c@{\hspace{0.8em}}
		c@{\hspace{0.8em}}c@{\hspace{0.8em}}c@{\hspace{0.8em}}
	}
	\toprule
	& \multicolumn{3}{c}{\textsc{SSR}} & \multicolumn{3}{c}{\textsc{SDR}} & \multicolumn{3}{c}{\textsc{Joint}} \\
	& \scriptsize{\textsc{RN50}}
	& \scriptsize{\textsc{IncV3}}
	& \scriptsize{\textsc{IncV3*}}
	& \scriptsize{\textsc{RN50}}
	& \scriptsize{\textsc{IncV3}}
	& \scriptsize{\textsc{IncV3*}}
	& \scriptsize{\textsc{RN50}}
	& \scriptsize{\textsc{IncV3}}
	& \scriptsize{\textsc{IncV3*}}
	\\
	\midrule
	\textsc{Chang~\etal\cite{chang2018explaining}}
		& 39.74 & 32.44 & 27.79
		& 44.21 & 45.53 & 37.81
		& 37.61 & 33.92 & 28.66 \\
	\textsc{Ours}
		& 17.54 & 18.18 & 17.37
		& 22.72 & 21.87 & 20.20
		& 16.95 & 15.21 & 14.06 \\
	\bottomrule

	\end{tabular}
	\label{tab:tv_evalutation}
\end{table}

\subsection{Test-time Augmentation of a Fine-grained Classifier}
\label{sub:improving_classification}

\begin{table}[t]
	\caption{Comparison of the classification performance using different test-time augmentation (TTA) methods on the CUB-200-2011 dataset. Besides the baselines (no TTA or ground truth bounding boxes), we also evaluated heuristic methods (random or center crop), content-aware methods (GradCam or a bird detector), and two different FIDO implementations (the original work of Chang~\etal and our proposed improvement). We report the accuracy (in \%).}
	\centering
	\begin{tabular}{
		l@{\hspace{2.5em}}
		c@{\hspace{1.7em}}
		c@{\hspace{1.7em}}
		c@{\hspace{1.7em}}
	}
	\toprule

	& \textsc{ResNet50}
	& \textsc{InceptionV3}
	& \textsc{InceptionV3*}

	\\
	\midrule
	\textsc{Baseline (BL)}
		& 82.78 & 79.86 & 90.32 \\
	\textsc{GT bounding boxes only}
		& 84.38 & 81.31 & 90.18 \\
	\textsc{BL + GT bounding boxes}
		& 84.55 & 81.65 & 90.70 \\
	\midrule
	\textsc{BL + random crop}
		& 83.41 & 80.45 & 89.99 \\
	\textsc{BL + center crop}
		& 83.83 & 81.07 & 90.16 \\
	\midrule
	\textsc{BL + gradient \cite{Korsch19_CSPARTS}}
		& 83.74 & 80.76 & 90.02 \\
	\textsc{BL + BirdYolo \cite{birdyolo}}
		& 83.81 & 81.12 & 90.39 \\
	\midrule
	\textsc{BL + FIDO \cite{chang2018explaining}}
		& 84.17 & 81.67 & 90.47 \\
	\textsc{BL + FIDO (ours)}
		& \textbf{84.67} & \textbf{81.77} & \textbf{90.51} \\
	\bottomrule

	\end{tabular}
	\label{tab:clf_evalutation}
\end{table}

Given a model fine-tuned on the CUB-200-2011 dataset, we evaluated in this experiment how we can use the estimated attribution masks to improve the classification performance of the model.
In addition to the prediction of the baseline models, we used different methods to extract one auxiliary crop from the original image and compute the prediction using this crop.
Then, we averaged the predictions and report the resulting accuracies in Table~\ref{tab:clf_evalutation}.
This way of classification improvement is widely used to different extent.
He~\etal\cite{resnet} or Szegedy~\etal\cite{inceptionv3}, for example, use ten or \num{144} crops in their work, respectively.
Hu~\etal\cite{hu2019see}, as another example, perform attention cropping to enhance the prediction of the classifier.
In our setup, we extracted a single crop using different methods, which we explain in the following.

\noindent\textbf{Ground-truth bounding boxes}:
We utilized the bounding box annotations of the CUB-200-2011 dataset.
First, we only used the crops identified by the bounding boxes.
Second, we combined the predictions from the cropped image and the original image, by averaging the predictions.

\noindent\textbf{Center and random crop}:
Following the motivation behind the crops used by He~\etal\cite{resnet} and Szegedy~\etal\cite{inceptionv3} that the object of interest is likely to be in the center, we cropped the center of the image.
Furthermore, we also extracted a random crop.
For both methods, we set the size of the crop to be \pcent{75} of the width and height of the original image.
These methods are content-agnostic and use only heuristics to estimate the region to crop.

\noindent\textbf{Gradient crop}:
As a first content-aware method, we computed the gradients \wrt the input image~\cite{simonyan2014deep}.
We utilized the pre-processing and thresholding of the gradient as presented by Korsch~\etal~\cite{Korsch19_CSPARTS}, estimated a bounding box around the resulting saliency map, and cropped the original image based on the estimated bounding box.

\noindent\textbf{BirdYOLO} is a YOLOv3~\cite{yolov3} detection model pre-trained on a bird detection dataset~\cite{birdyolo}.
For each image, we used the bounding box with the highest confidence score, extended it to a square, and cropped the original image accordingly.

\noindent\textbf{FIDO}:
Finally, we utilized the joint mask computed from the SSR and SDR masks of the FIDO algorithm as defined in Eq.~\ref{eq:joint_mask}.
On the one hand, we used the masks estimated by the original work of Chang~\etal~\cite{chang2018explaining}, and on the other hand, the masks estimated with our proposed improvements.

The results in Table~\ref{tab:clf_evalutation} show that compared to the baseline model the ground truth bounding boxes yield a higher classification accuracy, even if solely using the bounding box crops for classification.
Next, we can see that even such content-agnostic methods like center or random cropping can boost classification performance.
Similar improvements can be achieved by content-aware methods like gradients or a detection model.
We observed the most improvement with the FIDO algorithm, and finally with our proposed solution we achieved the best results that are comparable to using ground-truth bounding boxes.

\section{Conclusions}
\label{sec:conclusions}
In this paper, we proposed a simplified version of the concrete dropout (CD).
The CD is used in the fill-in of the dropout (FIDO) algorithm to sample a set of attribution masks based on an underlying parametrized distribution.
Using these masks, one can estimate how relevant a specific image pixel was for the classification decision.
The parameters of the distribution are optimized based on the classification score but the optimization process suffers from a high variance in the gradient computation if the original formulation of CD is used.
Our solution simplifies the sampling computations and results in more stable gradient estimations.
Our approach maintains the quality of the estimated masks while reducing computational effort due to smaller mini-batch sizes during the optimization process.
Furthermore, the resulting attribution masks contain fewer falsely attributed regions.
We also presented a way of using the estimated fine-grained attribution masks to enhance the classification decision.
Compared with other classification baselines, our solution produces the best result and even performs comparably to a setup where ground truth bounding boxes are used.

As an extension, our proposed single-crop TTA can be extended with a part-based approach to further boost the classification performance.
Alternatively, a repeated iterative estimation of the masks may be worth an investigation.

\bibliographystyle{splncs04}
\bibliography{069-main}

\end{document}